\documentclass[letterpaper,10pt,conference]{ieeeconf}
\IEEEoverridecommandlockouts       

\usepackage{amssymb}
\usepackage{amsmath}
\usepackage[english]{babel}
\usepackage{float}
\usepackage{graphicx}

\usepackage{mathtools}
\usepackage[linesnumbered,ruled,vlined]{algorithm2e}
\usepackage{filecontents}
\usepackage[noadjust]{cite}
\usepackage[font=footnotesize,labelfont=footnotesize]{subcaption}
\usepackage[font=footnotesize,labelfont=footnotesize]{caption}
\usepackage{booktabs}
\usepackage{comment}
\usepackage{authblk}
\usepackage{multirow}
\usepackage{xcolor}
\usepackage{hyperref}
\usepackage{xspace}
\usepackage{float}

\definecolor{darkblue}{rgb}{0.15,0.15,0.55}
\definecolor{lightgrey}{rgb}{0.75,0.75,0.75}

\providecommand{\codecomment}[1]{\textcolor{lightgrey}{\dotfill}\textcolor{darkblue}{//\,\textrm{#1}}}

\SetFuncArgSty{\textnormal}

\let\oldtwocolumn\twocolumn
\renewcommand\twocolumn[1][]{%
    \oldtwocolumn[{#1}{
        \begin{center}
        \begin{tabular}{cc}
            \includegraphics[width=0.43\textwidth]{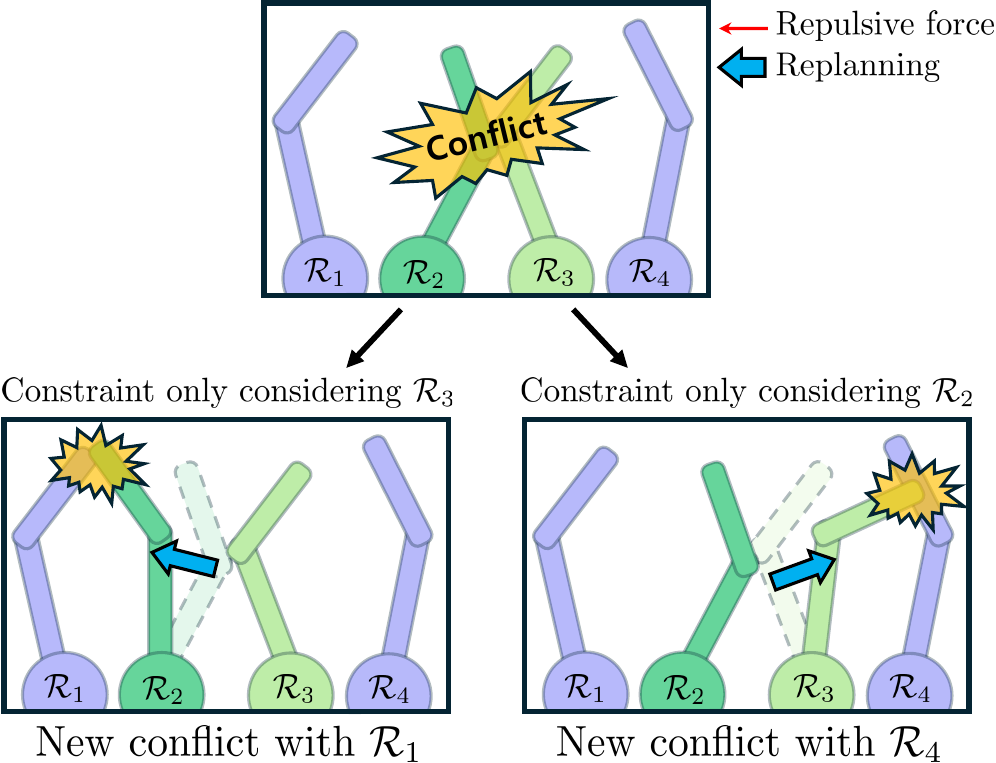} 
            \includegraphics[width=0.43\textwidth]{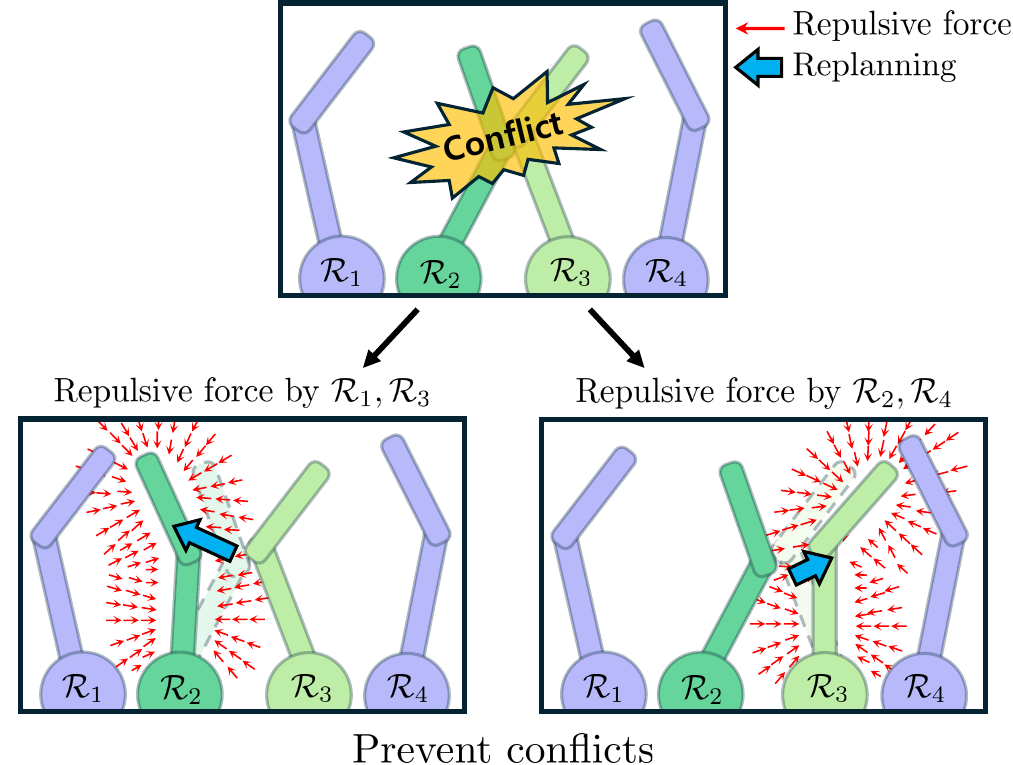} \\
            {\footnotesize (a) Replanned trajectories incurring subsequent conflicts} 
            {\footnotesize \qquad \qquad (b) Replanned with repulsive forces to avoid subsequent conflicts} \\
        \end{tabular}
        \end{center}\vspace{-10pt}
        \begin{center}
        \captionof{figure}{Examples of conflict trees expanded with CBS and our method, where the figure in each node represents a snapshot of robot trajectories. (a) Resolving the conflict between $\mathcal{R}_2$ and $\mathcal{R}_3$ by adding constraints can generate new conflicts, as the replanned trajectory does not consider entire robot trajectories. (b) Replanning with the repulsive trajectory modification promotes separation among trajectories which prevents subsequent conflicts.}
        \label{fig:illustration}
        \end{center}
        \vspace{-10pt}
    }]
}

\title{\LARGE \bf Repulsive Trajectory Modification and Conflict Resolution for\\Efficient Multi-Manipulator Motion Planning\vspace{-7pt}}

\author{Junhwa Hong$^1$, Beomjoon Lee$^1$, Woojin Lee$^2$, and Changjoo Nam$^{1,*}$\thanks{This work was supported by the National Research Foundation of Korea (NRF) grants funded by the Korea government (MSIT) (No. 2022R1C1C1008476 and No. RS-2024-00461583). $^{1}$Dept. of Electronic Engineering, Sogang University, Korea. $^{2}$Dept. of Artificial Intelligence, Sejong University, Korea. $^{*}$Corresponding author: {\tt\small cjnam@sogang.ac.kr}}}

\begin{document}
\maketitle
\begin{abstract}
We propose an efficient motion planning method designed to efficiently find collision-free trajectories for multiple manipulators. While multi-manipulator systems offer significant advantages, coordinating their motions is computationally challenging owing to the high dimensionality of their composite configuration space. Conflict-Based Search (CBS) addresses this by decoupling motion planning, but suffers from subsequent conflicts incurred by resolving existing conflicts, leading to an exponentially growing constraint tree of CBS. Our proposed method is based on repulsive trajectory modification within the two-level structure of CBS. Unlike conventional CBS variants, the low-level planner applies a gradient descent approach using an Artificial Potential Field. This field generates repulsive forces that guide the trajectory of the conflicting manipulator away from those of other robots. As a result, subsequent conflicts are less likely to occur. Additionally, we develop a strategy that, under a specific condition, directly attempts to find a conflict-free solution in a single step without growing the constraint tree. Through extensive tests including physical robot experiments, we demonstrate that our method consistently reduces the number of expanded nodes in the constraint tree, achieves a higher success rate, and finds a solution faster compared to Enhanced CBS and other state-of-the-art algorithms.

\end{abstract}
\vspace{-3pt}

\section{Introduction}
\vspace{-3pt}

Although multi-manipulator systems enable substantial gains through parallel execution, coordinating their motions is computationally challenging owing to the high dimensionality of the composite configuration space (C-space). Conflict-Based Search (CBS)~\cite{sharon2015conflict} and its variants~\cite{barer2014suboptimal,li2021eecbs} have been alternatives for improved efficiency, which operate in two levels. The high-level search performs a best-first search on a constraint tree (CT) whose nodes contain paths for each robot and constraints on individual robots. The low-level planner is invoked when a conflict is detected in a CT node to regenerate trajectories for the conflicting manipulators. CBS effectively reduces the size of the search tree compared to traditional state-based methods, whose search tree complexity is exponential in the number of robots.

Despite recent advancements of CBS and its variants, they remain prone to substantial search overhead, as its high-level search can grow exponentially with the number of conflicts. This arises from resolving conflicts through local, hard constraints on individual robots without considering the full trajectories of others. A trajectory adjusted to satisfy one constraint often collides with another, generating subsequent conflicts as shown in Fig.~\thefigure a. This issue becomes more frequent in multi-manipulator systems with fixed bases, where robots operate in close proximity and share overlapping workspaces. Each such conflict introduces new branches in the high-level search, increasing the number of motion planning calls and overall computation time. 

We propose a Multi-manipulator Motion Planning (MMP) method that addresses the frequent occurrence of new conflicts from replanning in CBS. Our method efficiently finds collision-free trajectories by \textit{repulsive trajectory modification} (repulsive modification in short). This approach modifies the trajectory using a gradient descent that moves along the repulsive force of an Artificial Potential Field (APF)~\cite{khatib1986real}, acting as a soft constraint defined between the manipulators. The trajectories tend to be separated, thereby reducing the likelihood of subsequent conflicts, as illustrated in Fig.~\thefigure b.

In addition to the modification in the low level, we propose an efficient strategy for the high-level search. For a node where an analysis reveals that all conflicts are attributable to a single manipulator, we execute the repulsive modification without sending the control to the low-level trajectory replanning. In many cases, this modification in the high level can resolve all conflicts in a single attempt, which can yield a conflict-free solution at an early stage.

Our key contributions include (i) the low-level strategy to promote separations between trajectories to prevent newly incurred conflicts while resolving existing conflicts, (ii) the high-level strategy to find an opportunity to resolve all conflicts at once, enabling an early termination of CBS. Lastly, (iii) we provide extensive experiments with up to eight high degree-of-freedom (DOF) manipulators in simulation and with physical robots.

\section{Related Work}
\vspace{-3pt}

Multi-Agent Path Finding (MAPF) coordinates robots to generate collision-free paths. Coupled methods like A$^*$ and RRT~\cite{lavalle2001randomized,kuffner2000rrt} search the joint state space but scale poorly, while decoupled methods such as Prioritized Planning (PP)~\cite{erdmann1987multiple} are efficient yet incomplete. CBS offers a hybrid solution by planning paths individually and resolving conflicts only when they arise. Its extension, Enhanced CBS (ECBS)~\cite{barer2014suboptimal}, accelerates the process with bounded suboptimality, making conflict-based approaches more scalable than exhaustive joint search.

Nevertheless, CBS can still be computationally expensive for multi-robot manipulation, where the motion planning needs to explore high-dimensional spaces and perform extensive collision checking. Various extensions of CBS have been proposed to address these limitations. CBS-MP~\cite{solis2021representation} modified constraints to prune the search space. xECBS~\cite{shaoul2024accelerating} accelerates low-level searches by reusing paths generated in the previous CT node. Generalized ECBS (Gen-ECBS)~\cite{shaoul2024unconstraining} enables the use of incomplete constraints while maintaining completeness, offering a trade-off between efficiency and completeness. However, these methods do not address the subsequent conflicts that arise from the decoupled strategy of resolving conflicts.

APF models the environment as a virtual force field, generating attraction to the goal and repulsion from obstacles. It is responsive in real time but suffers from limitations such as local minima, where forces balance in undesired configurations. To overcome the local minima problem, there has been active research into combining APF with search algorithms: APF was integrated into RRT$^*$-Connect to handle narrow passages~\cite{yan2025frrt}, and an APF-based sampling guide was proposed to improve the convergence of RRT$^*$~\cite{qureshi2016potential}. However, using APF as a sampling heuristic fails to address the multi-robot collision problem, where manipulators act as dynamic obstacles to each other. Additionally, reactive APF methods have been proposed for dual-arm robots, but these remain local avoidance strategies and are difficult to scale to multi-manipulator environments~\cite{zhang2021analysis}.

Our method incorporates APF-based trajectory modification into the low level of CBS to reduce cascading conflicts during replanning. At the high level, we add a strategy that enables the early termination of the search, resulting in faster planning times on average. Together, these improvements allow our method to efficiently find conflict-free solutions, scaling to eight high-DOF manipulators operating in heavily overlapped task spaces.

\section{Problem Definition}
\label{sec:prob}
\vspace{-3pt}

We consider the MMP problem of finding collision-free trajectories for $n$ manipulators $\mathcal{R}_i$ for $i \in \{1, 2, \dots, n\}$ from their start configuration $q^{\text{start}}_i$ to goal configuration $q^{\text{goal}}_i$. A configuration $q$ is in $\mathbb{R}^{d_i}$ where $d_i$ is the number of joints of $\mathcal{R}_i$.
Each trajectory of $\mathcal{R}_i$ is a sequence of tuples $\mathcal{T}_i = ((q_i^\text{start}, t^\text{start}_i), \dotsi,(q_i^\text{goal}, t^\text{goal}_i))$. 
Each joint angle and its corresponding velocity must be bounded by the joint limits and maximum velocity $v_\text{max}$. We assume that $\mathcal{R}_i$ occupies a volume $\mathcal{R}_i(q_i, t) \in \mathbb{R}^3$ in the task space with the configuration $q_i$ at $t$. 
Let $C_i \in \mathbb{R}^{d_i}$ be the free configuration space of $\mathcal{R}_i$ so $\mathcal{R}_i(q_i,t)$ is collision-free if $q_i \in C_i$ at $t$. A conflict between $\mathcal{T}_i$ and $\mathcal{T}_j$ ($i \neq j$) occurs if $\mathcal{R}_i(q_i, t) \cap \mathcal{R}_j(q_j, t) \ne \emptyset$ at an arbitrary $t$. 

In this setting, our goal is to find a solution to the MMP problem, which is a set of trajectories $\mathcal{T} = \{\mathcal{T}_1,\dotsi,\mathcal{T}_n\}$ such that for all $i, j \in \{1, \dots, n\}$, the following conditions hold: 
(i) $\forall (q, t) \in \mathcal{T}_i : q \in C_i$, and 
(ii) if $i\neq j$, then $\forall t \in [0, T] : \mathcal{R}_i(q_i, t) \cap \mathcal{R}_j(q_j, t) = \emptyset$, where $T = \max_{k \in \{1, \dots, n\}} (t_k^{\text{goal}})$.\footnote{These two conditions ensure that all trajectories do not have collisions with any static obstacle and robot.}

\section{Preliminaries} 
\vspace{-3pt}

We briefly introduce ECBS and APF, which form the basis of our method.

\subsection{Enhanced CBS}
\vspace{-3pt}

In the CT of CBS, each node $N$ contains a set of trajectories $N.\mathcal{T} = \{\mathcal{T}_1, \dots, \mathcal{T}_n\}$, each $\mathcal{T}_i$ being an individually optimal trajectory for $\mathcal{R}_i$, and a cost defined as the sum of their lengths. CBS expands nodes in a best-first manner from a priority queue (OPEN). When a conflict between $\mathcal{T}_i$ and $\mathcal{T}_j$ is detected, $N$ is branched into two child nodes, each with a constraint forbidding the conflict for one robot, and the corresponding low-level replanning is performed.

ECBS is a bounded suboptimal variant of CBS. It maintains a focal list which is a subset of OPEN:
$$\textnormal{FOCAL} \coloneqq \left\{ N \in \textnormal{OPEN} \mid N.cost \le w \cdot \min_{N' \in \textnormal{OPEN}} lb(N') \right\}$$
where $lb(N') = \sum_{i=1}^n lb(\mathcal{T}_i)$ and $lb(\mathcal{T}_i)$ is a lower bound on the cost of $\mathcal{T}_i$. Nodes in FOCAL are prioritized by the number of conflicts. ECBS guarantees a solution with cost at most $w \cdot C^*$, where $C^*$ is the optimal cost and $w \geq 1$ is the suboptimality factor.

\subsection{Artificial Potential Fields}
\vspace{-3pt}

APF models the environment of a robot as a virtual force field with attractive forces toward the goal and repulsive forces from obstacles. The repulsive potential is defined between two points $p$ and $o$
\begin{equation}
U_{\text{rep}}(p, o) =
\begin{cases}
    \frac{1}{2} k_{\text{rep}} \left( \frac{1}{d(p, o)} - \frac{1}{d_0} \right)^2 & \text{if } d(p, o) \le d_0 \\
    0 & \text{otherwise}
\end{cases}
\end{equation}
where $k_\textnormal{rep}$ is the repulsive gain, $d(q, o)$ is the minimum Euclidean distance between $p$  and $o$, and $d_0$ is a radius of influence, defining the range at which the repulsive force becomes active.  In the context of collision avoidance, $p$ and $o$ are the points representing a robot and its nearest obstacle, respectively. The resulting repulsive force, $F_{\text{rep}}$, is derived from the negative gradient of this potential:
\begin{equation}
\label{eq:F_rep}
F_{\text{rep}}(p, o) = k_{\text{rep}} \left( \frac{1}{d(p, o)} - \frac{1}{d_0} \right) \frac{1}{d(p, o)^2} \nabla d(p, o)
\end{equation}
where $\nabla d(p, o)$ is a unit vector pointing from $o$ toward $p$. 

\section{The Proposed Method: APF-ECBS}
\vspace{-3pt}

In our proposed method APF-ECBS, the low level generates trajectories, while the high level detects conflicts in the CT, imposes new constraints, and iteratively invokes the low level to replan and apply the repulsive modification until a conflict-free solution is found.

\subsection{The Low Level of APF-ECBS}
\vspace{-3pt}

The low level generates and modifies trajectories of individual manipulators, using a sampling-based planner. This trajectory is then refined through repulsive modification, which incorporates the trajectories of other robots to reduce the likelihood of further conflicts.

\subsubsection{APF among multiple manipulators}
We first synchronize all $n$ trajectories for further refinements as they are planned individually. Through interpolation, they are synchronized with a common time interval $\Delta t$.   For any $t^\text{start} < t^k < t^\text{goal}$ where $t_k$ represents the $k$-th time step in a trajectory, all manipulators are considered to compute the repulsive forces. While the forces can be more accurately computed with the exact geometry of the robot, we approximate the robot geometry using simple collision bodies (e.g., spheres, cylinders) to reduce computational burdens. We denote the set of collision bodies of $\mathcal{R}_i$ at $t_k$ by $\mathcal{S}_i(q^k_i,t^k)$.

The repulsive force between two points $p_i \in \mathcal{S}_i(q^k_i,t^k)$ and $p_j \in \mathcal{S}_j(q^k_i,t^k)$ is denoted as ${F}(p_i, p_j)$ and computed by~\eqref{eq:F_rep}. Note that we compute the force between a pair of points, one from $\mathcal{R}_i$ and the other from $\mathcal{R}_j$. In our implementation, we model the collision bodies using spheres (as shown in Fig.~\ref{fig:fig_3_b}) so use their centroids to compute the force. The repulsive force ${F}(p_i,p_j)$ in the task space is mapped to the joint space using the Jacobian transpose, $J^\text{T}_{p_i}(q^k_i)$, which is computed at $p_i$. The resulting $F_\text{joint}$ on $\mathcal{R}_i$ at $t^k$, is the summation of these individual joint space force over all interacting point pairs in $\mathcal{S}_i(q^k_i,t^k)$:

{\small
\begin{equation}
\label{eq:total_repulsive_torque}
F_{\text{joint}}(t^k) = \sum_{j} \sum_{p_i \in S_i(q^k_i,t^k)} \sum_{p_j \in S_j(q^k_j,t^k)} J^\text{T}_{p_i}(q^k_i) {F}(p_i,p_j)
\end{equation}}
which is repeated for every $t^k$, yielding a repulsive force
\begin{equation}
\boldsymbol{\Gamma}_{\text{joint}} = \left( {F}_{\text{joint}}(t^\text{start}), \dotsi, {F}_{\text{joint}}(t^\text{goal})\right).
\end{equation}

\subsubsection{Repulsive trajectory modification}
The repulsive modification (Alg.~\ref{alg:modifymotion}) on $\mathcal{T}_i$ of $\mathcal{R}_i$ is done through a gradient descent process, where the trajectory is incrementally adjusted in a direction that decreases potential collisions:
\begin{equation}
\label{eq:update}
\mathcal{T}^\prime_i = \mathcal{T}_i + \alpha \boldsymbol{\Gamma}_\textnormal{joint}
\end{equation}
where the addition is performed element-wise and $\alpha$ is a step size.\footnote{The choice of initial $\alpha$ presents a trade-off between convergence speed and stability. Large initial $\alpha$ can lead to faster convergence, but risks overshooting the solution and causing instability, while small $\alpha$ ensures stability at the cost of slower convergence.} We note that ${F}_{\text{joint}}(t^\text{start})$ and ${F}_{\text{joint}}(t^\text{goal})$ are set to zero to maintain the start and the goal configurations of the trajectory. The repulsive modification (i.e., gradient descent update) is applied iteratively until the trajectory collides with a static obstacle or a predefined maximum number of iterations is reached. During the iteration, a collision-free trajectory may not satisfy the high-level constraint imposed in the CT. If so, the modification continues as subsequent iterations would be able to guide the trajectory to a new trajectory that satisfies the constraint.


%

Notice that this modification may introduce trajectory segments that violate $v_\text{max}$ by increasing the path length while maintaining the same time interval, which can result in velocities exceeding the allowed maximum. A synchronized time-scaling is applied to those segments, uniformly stretching their time parameterization so that no segment exceeds $v_\text{max}$.


\subsection{The High Level of APF-ECBS} 
\label{sec:conflict_resolution} 
\vspace{-3pt}

The high level of APF-ECBS follows ECBS, expanding nodes from FOCAL in order of conflict count. If a node has zero conflict, the corresponding set of trajectories $\mathcal{T}$ is a solution and the search terminates. Otherwise, the conflict resolution process (Alg.~\ref{alg:cbs_repulsive}, lines~\ref{conflict_resol}--\ref{conflict_resol_done}) generates two child nodes. Each child inherits the parent’s solution and constraints, while adding a new constraint for one of the robots involved. In the left child, $(\mathcal{R}_i,q_i,t^k)$ forbids $\mathcal{R}_i$ from occupying configuration $q_i$ at $t^k$, and the low-level planner replans $\mathcal{R}_i$ accordingly. The right child is constructed analogously with $(\mathcal{R}_j,q_j,t^k)$. After this replanning, conflict checking is performed by iterating through every pair of trajectories $\mathcal{T}_i$ and $\mathcal{T}_j$ at each time step. Both child nodes are then added to the open list for subsequent search.

\begin{figure}
\vspace{-5pt}
\captionsetup{skip=0pt}
    \centering
    \begin{subfigure}{0.5\textwidth}
        \captionsetup{skip=0pt}
        \centering
	\includegraphics[width=0.53\textwidth]{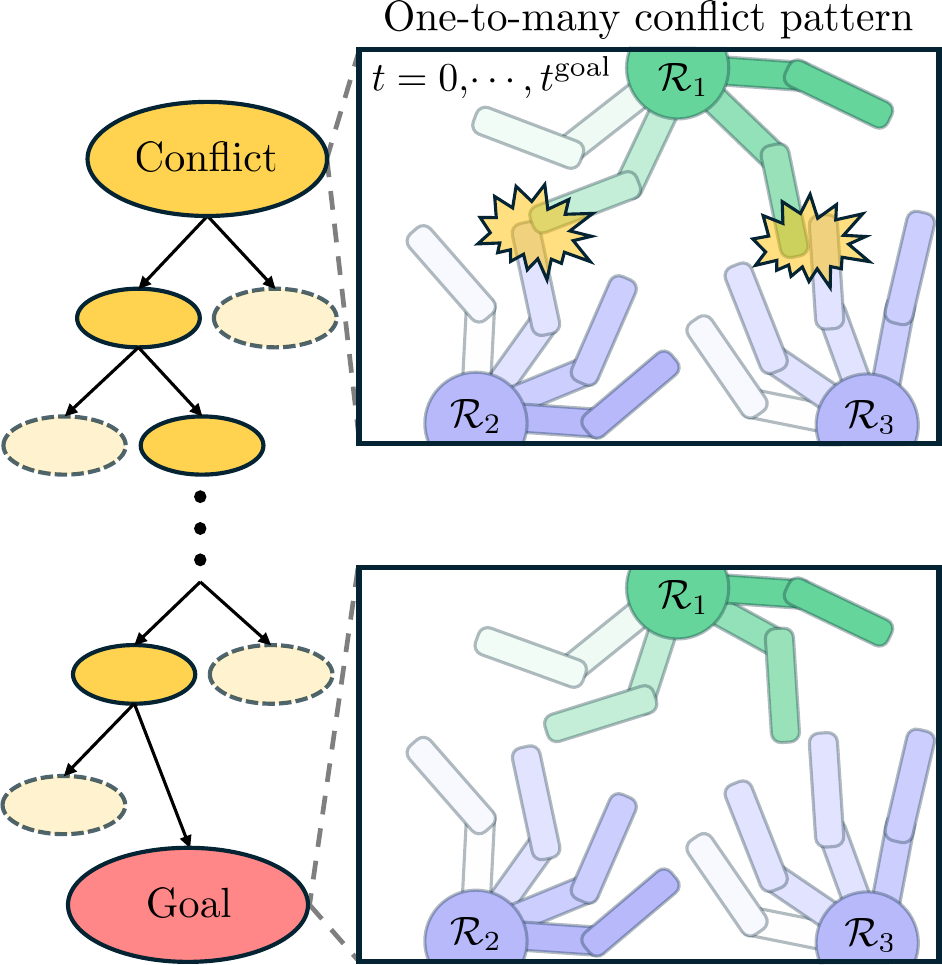}
	\caption{ECBS expanding the CT until finding a solution}
    \label{fig:fast_ecbs}
  \end{subfigure}\\
  \vspace{2pt}
  \begin{subfigure}{0.5\textwidth}
        \captionsetup{skip=0pt}
        \centering
	\includegraphics[width=0.53\textwidth]{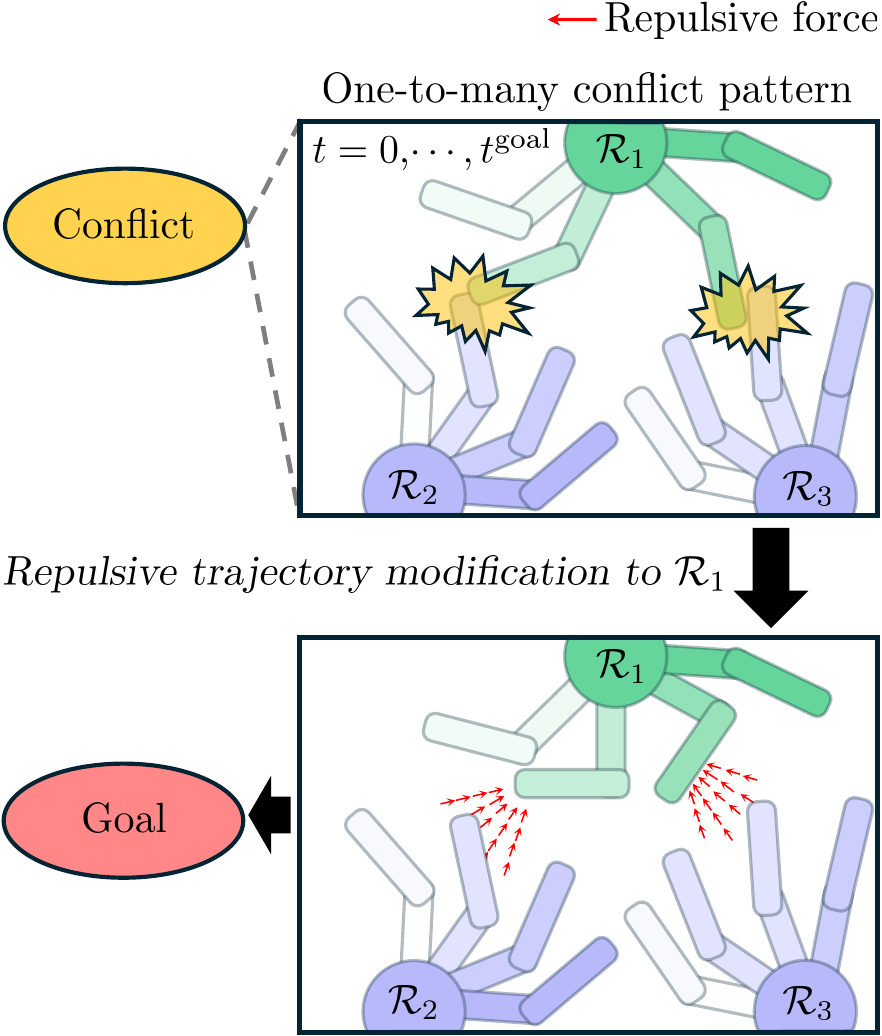}
    \caption{Our fast-track directly resolving conflicts without further search}
    \label{fig:fast_ours}
  \end{subfigure}
  \caption{The case of one-to-many conflicts. In the example, $\mathcal{R}_1$ is responsible for all conflicts at a given node. (a) As the low-level of ECBS does not particularly detect and handle this type of conflicts, it expands the CT nodes until a solution is found. (b) In our method, we detect this case and then attempt to resolves all conflicts at once by applying the repulsive modification to the trajectory of $\mathcal{R}_1$, so directly reaching to a solution.}
  \label{fig:fasttrack}
  \vspace{-16pt}
\end{figure}

This search is computationally expensive because each branching requires an intensive low-level replanning. Before proceeding with the search, we first attempt a fast-track strategy (lines~\ref{fast_track}--\ref{fast_track_done} in Alg.~\ref{alg:cbs_repulsive}). This strategy addresses a one-to-many conflict pattern in which a single manipulator termed the critical robot ($\mathcal{R}_c$), is involved in all conflicts of the node. In this case, resolving the conflicts related only to $\mathcal{T}_c$ eliminates all conflicts, since the other robots do not interfere with one another. Fig.~\ref{fig:fasttrack} shows an example of the one-to-many conflicts handled by ECBS and our fast-track strategy.

The case of one-to-many conflicts can be found easily by examining the set of unique conflict pairs, where each pair $(\mathcal{R}_i, \mathcal{R}_j)$ represents a conflict between the two robots. For example, if the set of conflict pairs is $\{(\mathcal{R}_1,\mathcal{R}_2),(\mathcal{R}_1,\mathcal{R}_3)\}$, it is clear that only $\mathcal{R}_1$ is present in every pair. Thus, $\mathcal{R}_c = \mathcal{R}_1$. If $\mathcal{R}_c$ is found, the repulsive modification is performed on $\mathcal{R}_c$ only. The resulting trajectory of $\mathcal{R}_c$ is likely to be separated from other trajectories, giving a high probability of resolving all conflicts. 

If all conflicts are resolved by the fast-track strategy, Alg.~\ref{alg:cbs_repulsive} updates the trajectory set, cost, and conflict list of the current node $N$. Then, $N$ with the conflict-free trajectories will be re-evaluated with all other candidates in OPEN rather than terminating the search immediately. This ensures that any potential node in OPEN with lower-cost conflict-free trajectories is not ignored. If some conflicts remains even after the fast-track strategy, these modifications are discarded, and the method proceeds to the conflict resolution process described above using the original $N$.
\vspace{5pt}

\begin{algorithm}
\small{
\SetAlgoNoEnd
\caption{\textsc{APF-ECBS}}
\label{alg:cbs_repulsive}
\SetKwInput{KwIn}{Input}

\KwIn{\parbox[t]{0.84\algowidth}{
    Number of robots $n$, Sets of initial and goal configurations of all $n$ robots $q^\text{start} = \{q_1^\text{start},\dots,q_n^\text{start}\}$ and $q^\text{goal} = \{q_1^\text{goal},\dots,q_n^\text{goal}\}$
}}

\KwOut{Collision-free trajectories of $n$ robots $\mathcal{T}$}

\vspace{1em} 
\BlankLine
\SetKwProg{Fn}{Procedure}{}{}

\Fn{\textsc{InitRootNode}$(n, q^\textnormal{start}, q^\textnormal{goal})$}{
    $\textnormal{RootNode}.\mathcal{C} \leftarrow \emptyset$, $\textnormal{RootNode}.\mathcal{T} \leftarrow \emptyset$, $\textnormal{RootNode}.\Omega\leftarrow \emptyset$,  $\mathcal{T}_\textnormal{all} \leftarrow \emptyset$\\
    \For{$i = 1$ \KwTo $n$}{
        $\mathcal{T}_i \leftarrow \textsc{MotionPlanner}(q_i^\textnormal{start}, q_i^\textnormal{goal})$\\
        $\mathcal{T}_\textnormal{all} \leftarrow \mathcal{T}_\textnormal{all} \cup \mathcal{T}_i$
    }
    
    $\textnormal{RootNode}.\mathcal{T} \leftarrow \mathcal{T}_\textnormal{all}$\\
    $\textnormal{RootNode}.\text{cost} \leftarrow \textsc{GetCost}(\textnormal{RootNode}.\mathcal{T})$\\
    $\textnormal{RootNode}.\Omega \leftarrow \textsc{GetConflicts}(\textnormal{RootNode}.\mathcal{T})$\\
    $\textnormal{RootNode}.\mu \leftarrow \textnormal{FALSE}$\textnormal{\codecomment{Fast-track not yet attempted}}\\
    \Return{\textnormal{RootNode}}
}
\BlankLine
\Fn{\textsc{Plan}\textnormal{(}$n, q^\textnormal{start}, q^\textnormal{goal}$\textnormal{)}}{
    $\textnormal{OPEN} \leftarrow \emptyset$\\
    \textnormal{OPEN}.\text{\textsc{insert}(\textsc{InitRootNode}\textnormal{(}$n, q^\textnormal{start}, q^\textnormal{goal}$\textnormal{)}\textnormal{)}}\\
    \While{$\textnormal{OPEN} \neq \emptyset$}{
        $\textnormal{FOCAL}\gets$ $\{N\in\textnormal{OPEN}\mid N.\text{cost}\le w\cdot{\min\limits_{N^\prime \in \textnormal{OPEN}}}\text{lb}(N^\prime)\}$
        \textnormal{\codecomment{where $lb(N^\prime) \coloneqq \sum_{i=1}^{n}lb(\mathcal{T}_i)$ and $lb(\mathcal{T}_i)$ is a lower bound on the cost of $\mathcal{T}_i$}}
        
        $N\leftarrow\textsc{GetBestNode(FOCAL)}$\\
        \If{$N.\Omega=\emptyset$}{
            $\mathcal{T} \leftarrow \textsc{ScaleTime}(N.\mathcal{T})$\\
            \Return {$\mathcal{T}$}\\}
        $c\gets\textsc{FindCriticalRobot}(N)$\label{fast_track}\textnormal{\codecomment{The conditions for the fast-track are detailed in Section~\ref{sec:conflict_resolution}}}\\
        \If{$c\textnormal{ is not null} \textnormal{\textbf{ and }} N.\mu = \textnormal{FALSE}$}{
            $N.\mu \leftarrow \textnormal{TRUE}$\\
            $M \gets \textsc{CopyCTNode}(N)$\\
            $M.\mathcal{T}_c \gets \textsc{ModifyMotion}(M.\mathcal{T}_c, M.\mathcal{T} \setminus M.\mathcal{T}_c, M.\mathcal{C})$\\
            $M.\Omega\gets\textsc{GetConflicts}(M.\mathcal{T})$\\
            \If{$M.\Omega=\emptyset$ \label{fast_track_success1}}{
                $N.\mathcal{T}\gets M.\mathcal{T}$\\
                $N.cost\gets\textsc{GetCost}(M.\mathcal{T})$\\
                $N.\Omega\gets M.\Omega$\\
                
                \textnormal{\textbf{continue}\codecomment{Skip branching. The modified node $N$ will be re-evaluated in \text{FOCAL} in the next iteration}}}\label{fast_track_done}
            
            }
        \textnormal{OPEN}.\textsc{remove}($N$)\\
        \For{$i \in \textsc{FirstConflictPair}(N)$}{ \label{conflict_resol}
            $N^\prime\leftarrow\textsc{CopyCTNode}(N)$\\
            $NewConstraint\leftarrow \textsc{CreateConstraint}(N.\Omega)$\\
            $N^\prime.\mathcal{C}\leftarrow N.\mathcal{C} \cup NewConstraint$\\
            $\mathcal{T}_i^\textnormal{new} \gets \textsc{MotionPlanner}(q_i^\text{start}, q_i^\text{goal}, N^\prime.\mathcal{C})$\\
            $N^\prime.\mathcal{T}_i \gets \textsc{ModifyMotion}(\mathcal{T}_i^\textnormal{new},N^\prime\mathcal{.T} \setminus N\mathcal{.T}_{i}, N^\prime.\mathcal{C})$\\
            $N^\prime.cost \gets \textsc{GetCost}(N^\prime.\mathcal{T})$\\
            $N^\prime.\Omega \gets \textsc{GetConflicts}(N^\prime.\mathcal{T})$\\
            $N^\prime.\mu \leftarrow \textnormal{FALSE}$\\
            \textnormal{OPEN}.\textsc{insert($N^\prime$)}\label{conflict_resol_done} 
        }
    }
    \Return{$\emptyset$}}
}
\end{algorithm}

\begin{algorithm}
\small{
\SetAlgoNoEnd
\caption{\textsc{ModifyMotion}}
\label{alg:modifymotion}

\SetKwInput{KwIn}{Input}

\KwIn{
    Trajectory to be modified $\mathcal{T}_i$, Trajectories of all robots $\mathcal{T}_{\textnormal{all}}$, Constraints $\mathcal{C}$, Step size $\alpha$
}
\KwOut{Updated collision-free trajectory $\mathcal{T}_{\textnormal{safe}}$}

\BlankLine

$\mathcal{T}_{\textnormal{safe}} \leftarrow \mathcal{T}_i$ \\
$\mathcal{T}_{\textnormal{current}} \leftarrow \mathcal{T}_i$\\

\For{$k = 1$ \KwTo maxIter}{
    $\boldsymbol{\Gamma}_{\textnormal{joint}} \leftarrow \text{\textsc{ComputeRepulsiveForce}}(\mathcal{T}_{\textnormal{current}}, \mathcal{T}_{\textnormal{all}} \setminus \mathcal{T}_i)$
    
    $\Delta \mathcal{T} \leftarrow \alpha \boldsymbol{\Gamma}_{\textnormal{joint}}$\\
    $\Delta \mathcal{T}_{\textnormal{start}} \leftarrow 0$, $\Delta \mathcal{T}_{\textnormal{end}} \leftarrow 0$\\
    $\mathcal{T}_{\textnormal{next}} \leftarrow \mathcal{T}_{\textnormal{current}} + \Delta \mathcal{T}$
    
    \If{\textnormal{\textbf{not}} \textsc{isCollisionFree}($\mathcal{T}_{\textnormal{next}}$)}{
        \textbf{break} \codecomment{Terminate on collision, the last $\mathcal{T}_{\textnormal{safe}}$ will be returned}
    }
    
    $\mathcal{T}_{\textnormal{current}} \leftarrow \mathcal{T}_{\textnormal{next}}$
    
    \If{\textnormal{\textbf{not}} \textsc{ViolatesConstraint}($\mathcal{T}_{\textnormal{current}}, \mathcal{C}$)}{
        $\mathcal{T}_{\textnormal{safe}} \leftarrow \mathcal{T}_{\textnormal{current}}$
    }
}
\Return $\mathcal{T}_{\textnormal{safe}}$
}
\end{algorithm}

\section{Experiments}
\vspace{-3pt}

\begin{figure*}
    \captionsetup{skip=0pt}
    \centering
    \begin{subfigure}{0.68\textwidth} 
        \captionsetup{skip=0pt}
        \centering
        \includegraphics[width=0.9\textwidth]{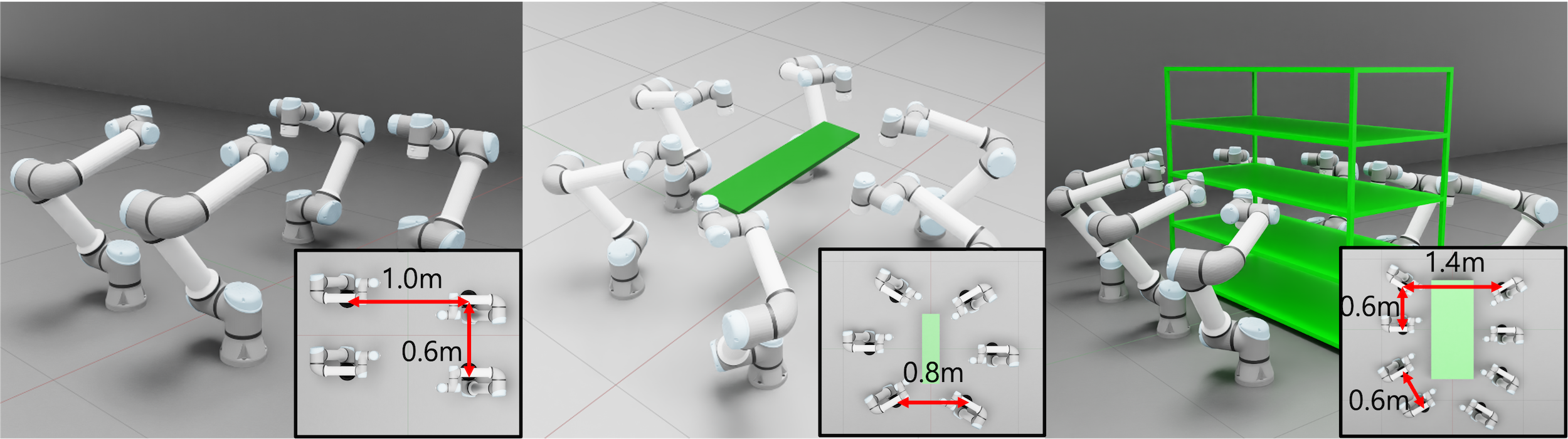}
        \caption{}
        \label{fig:fig_3_a}
    \end{subfigure}\quad
    \begin{subfigure}{0.17\textwidth}
        \captionsetup{skip=0pt}
        \centering
        \includegraphics[width=0.9\textwidth]{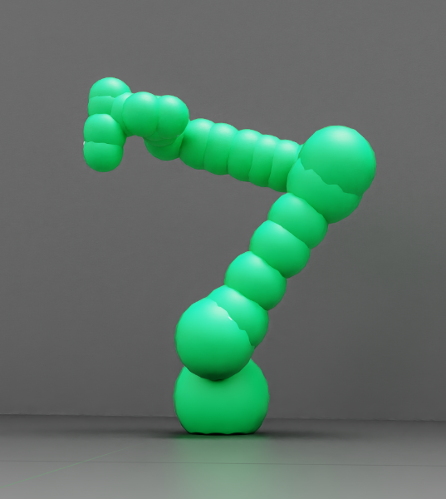}
        \caption{}
        \label{fig:fig_3_b}
    \end{subfigure}
    \caption{(a) Simulated environments with 4, 6, and 8 manipulators. (b) A sphere-based collision body.}
    \label{fig:fig_3_all}
    \vspace{-10pt}
\end{figure*}

We evaluate the computational efficiency of the proposed method against CBS, ECBS, and the state-of-the-art Gen-ECBS. We test scalability with up to eight manipulators in shared workspaces with closely located goals and demonstrate feasibility on four 6-DoF physical manipulators.

\subsection{Setup}
\vspace{-3pt}

For comparison, we designed three simulated environments using Isaac Sim~\cite{NVIDIA_Isaac_Sim} with four, six, and eight manipulators (Fig.~\ref{fig:fig_3_a}) where the distances between robot bases range from $0.6$m to $1.4$m, creating densely packed workspaces with frequent conflicts. In each environment, we generated $50$ random instances by uniformly sampling $q^\text{start}$ and $q^\text{goal}$ from the common task space of robots.
 
We measure various performance metrics which are (i) the number of expanded nodes in the high-level search, (ii) success rate, (iii) planning time, (iv) makespan, and (v) cost. The number of expanded nodes serves the most direct measure as our method is designed to reduce the efforts in the high-level search by reducing subsequent conflicts through the repulsive modification in the low level. The success rate is the percentage of the problem instances solved within a time limit of $1800$ seconds.\footnote{The implementation of all compared methods including ours is done in Python. We expect much faster execution if they are implemented in C++.} The planning time (of successful instances) measures the computation time until a solution is found. Finally, the quality of the solution is assessed by its makespan, which is the time until the last manipulator reaches its goal as well as the cost, defined as the sum of all joint movements of all robots (in radians).

To ensure a fair comparison, all methods used an identical low-level planner, SI-RRT$^*$~\cite{sim2024safe} which finds high-quality trajectories in dynamic environments. Collision-checking was done using the CUDA Accelerated Robot Library (cuRobo)~\cite{sundaralingam2023curobo}. The collision bodies of a robot is with $28$ spheres as shown in Fig.~\ref{fig:fig_3_b}. The parameters were set as follows: $\alpha = 10^{-7}$, $k_\textnormal{rep} =  0.05$, and $v_\textnormal{max} = 0.8\textnormal{rad/s}$. The suboptimality factor, $\omega$, was set to $1.5$ for ECBS, Gen-ECBS, and APF-ECBS. All experiments were done with a system with an AMD 5700X 3.8GHz CPU, 32GB RAM, and Python3.  

\subsection{Results}
\vspace{-3pt}

\begin{figure}[h!]
    \captionsetup{skip=0pt}
    \centering
    \includegraphics[width=0.43\textwidth]{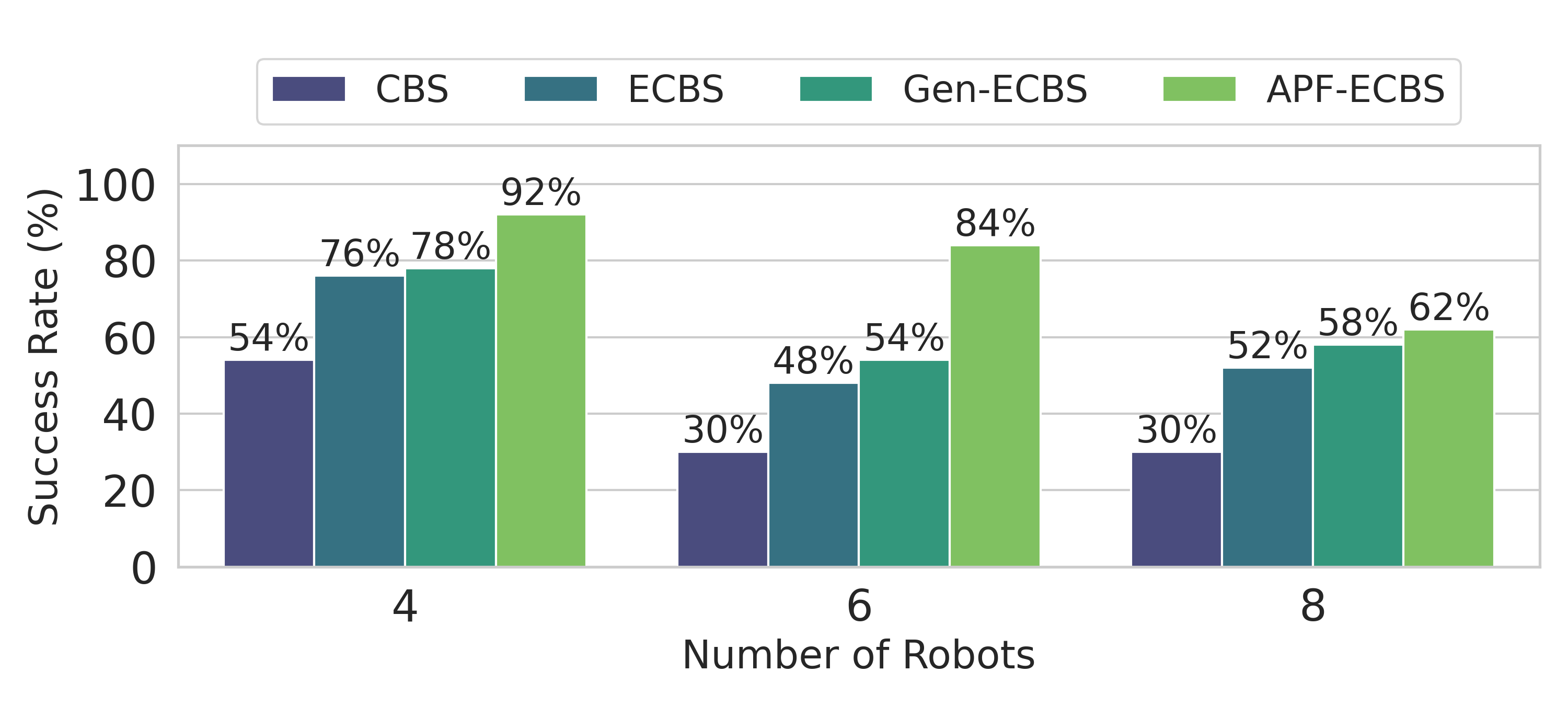}
    \caption{The success rates of the compared methods}
    \label{fig:success_rate}
    \vspace{-10pt}
\end{figure}

Fig.~\ref{fig:success_rate} shows that APF-ECBS achieves the highest success rate up to $92$\% in all environments while the compared methods reach $78$\% at most. As shown in Table~\ref{tab:all_results}, the number of expanded nodes for APF-ECBS is significantly lower than the compared methods. This confirms that our repulsive modification effectively prevents conflicts during search. 

\begin{figure*}[h!]
    \captionsetup{skip=0pt}
    \centering
    \includegraphics[width=0.95\textwidth]{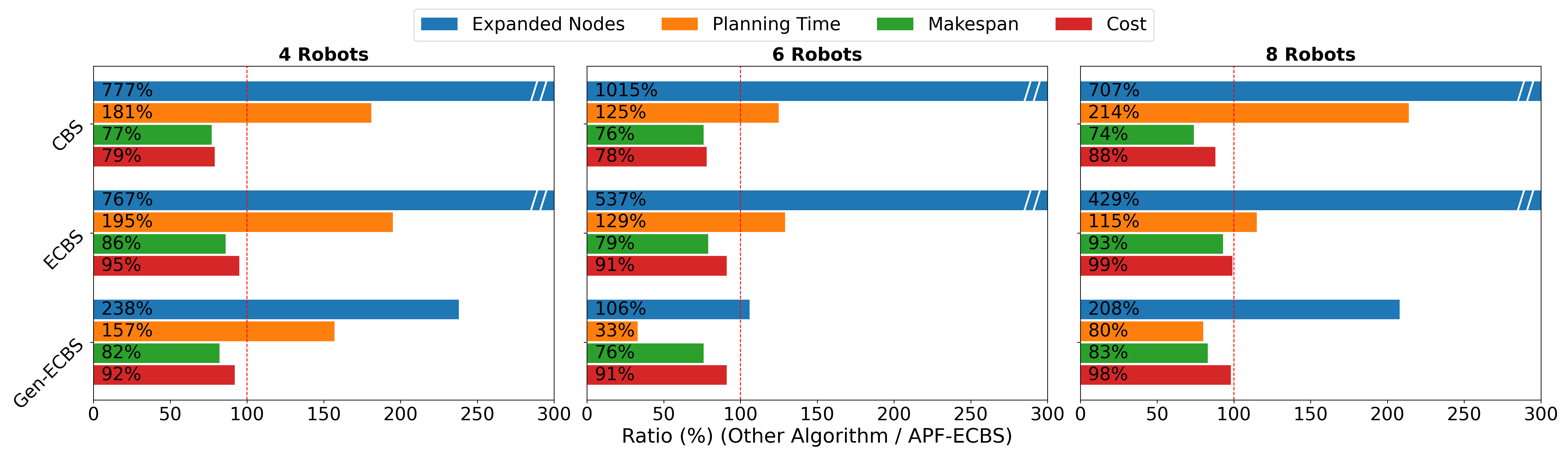}
    \caption{Pairwise comparisons between APF-ECBS and the other methods}
    \label{fig:pairwise}
    \vspace{-10pt}
\end{figure*}

Fig.~\ref{fig:pairwise} shows pairwise comparisons on instances solved by all methods, with performance measured relative to APF-ECBS ($100$\% indicating identical performance). For the number of expanded nodes (blue), all baselines expand more nodes in every environment, with ratios ranging from $208$\% to $1015$\%, except for Gen-ECBS in the 6-robot case. For planning time, APF-ECBS is consistently faster than CBS and ECBS. Although Gen-ECBS is slightly faster on the commonly solved instances in the 6- and 8-robot scenarios, Fig.~\ref{fig:success_rate} shows that APF-ECBS solves harder problems that Gen-ECBS fails to address. This search efficiency, however, comes at the expense of solution quality: APF-ECBS produces trajectories with larger makespan and cost than the baselines. The results highlight a clear trade-off: our method sacrifices path quality for substantial gains in speed and robustness, making it well suited for time-critical or difficult problems.

\begin{table*}
\captionsetup{skip=0pt}
\caption{Comparison result. Planning time and makespan are measured in seconds, cost is measured in radians, and success rate is given in percentage.}
\label{tab:all_results}
\centering
\resizebox{0.75\textwidth}{!}{%
\begin{tabular}{|c|l||c|c|c|c|c|}
\hline
\text{\#Robots} & \text{Method} & \text{\#Expanded nodes} & \text{Planning time} & \text{Makespan} & \text{Cost} & \text{Success rate} \\
\hline
\hline
\multirow{5}{*}{\textbf{4}} 
& CBS & 269.58 (246.36) & 337.71 (439.71) & \textbf{10.71} \textbf{(3.64)} & \textbf{45.48} \textbf{(12.65)} & 54 \\
& ECBS & 195.87 (266.85) & 363.88 (490.13) & 12.03 (3.87) & 54.25 (15.97) & 76 \\
& Gen-ECBS & 60.72 (71.79) & 293.16 (333.86) & 11.42 (3.05) & 52.91 (13.18) &  78\\
& APF-ECBS-NF & 35.00  (62.80) & 201.06 (316.99) & 14.33 (3.56) & 58.85 (12.71) & \textbf{96} \\
& APF-ECBS & \textbf{25.54 (37.78)}  & \textbf{186.71 (261.83)} & 13.94 (3.76) & 57.49 (14.34) & 92\\
\hline
\multirow{5}{*}{\textbf{6}} 
& CBS & 163.13 (145.62)  & 627.28 (517.04) & \textbf{12.68} \textbf{(3.03)} & \textbf{73.20} \textbf{(9.17)} & 30\\
& ECBS & 122.08 (110.26) & 617.40 (571.67) & 13.81 (3.15) & 85.97 (12.65) & 48 \\
& Gen-ECBS & 82.07 (55.48) & 667.99 (482.93) & 15.32 (4.60) & 92.29 (17.59) &  54\\
& APF-ECBS-NF & 44.9 (39.33) & 566.42 (494.46) & 17.85 (3.73) & 97.67 (15.38) &  82\\
& APF-ECBS & \textbf{37.24} \textbf{(32.28)} & \textbf{517.69} \textbf{(404.21)} & 17.80 (3.57) & 97.49 (17.36) &   \textbf{84}\\
\hline
\multirow{5}{*}{\textbf{8}} 
& CBS & 60.87 (40.67)  & 633.71 (330.65) & \textbf{18.29} \textbf{(4.29)} & \textbf{132.69} \textbf{(23.09)} & 30\\
& ECBS & 94.15 (72.63) & 745.29 (477.66) & 22.00 (7.13) & 149.87 (24.57) &  52\\
& Gen-ECBS & 68.59 (52.79)  & 796.55 (520.60) & 22.21 (8.06) & 153.77 (28.89) &  58\\
& APF-ECBS-NF & 34.03 (28.20) & 783.48 (488.18) & 23.61 (5.27) & 154.35 (23.48) &  \textbf{64}\\
& APF-ECBS & \textbf{21.94} \textbf{(19.39)} & \textbf{590.37} \textbf{(431.28)} & 24.38 (7.09) & 151.92 (28.33) &  62\\
\hline
\end{tabular}}
\vspace{-10pt}
\end{table*}

We also performed an ablation study to evaluate the impact of the fast-track strategy by comparing APF-ECBS algorithm against a variant without it, termed APF-ECBS-NF (Non-Fast-track). As shown in Table~\ref{tab:all_results}, APF-ECBS reduces the number of expanded nodes up to $39.81$\% and the planning time up to $24.65$\%  compared to the NF version, indicating that the fast-track strategy is effective in efficiently resolving simple conflict patterns without expanding CT. The success rates do not show meaningful differences, as they vary only by one or two instances. The results also highlight the fundamental effectiveness of our core contribution: the low-level repulsive modification. Even without the fast-track heuristic, the APF-ECBS-NF variant on its own significantly outperforms all baseline methods in terms of the number of expanded nodes. This demonstrates that our low-level modification is the primary reason for the reduced search effort, as it successfully prevents a large number of high-level conflicts.

\subsection{Physical-robot Demonstration}
\vspace{-3pt}

\begin{figure}[h!]
    \centering
    \includegraphics[height=4cm]{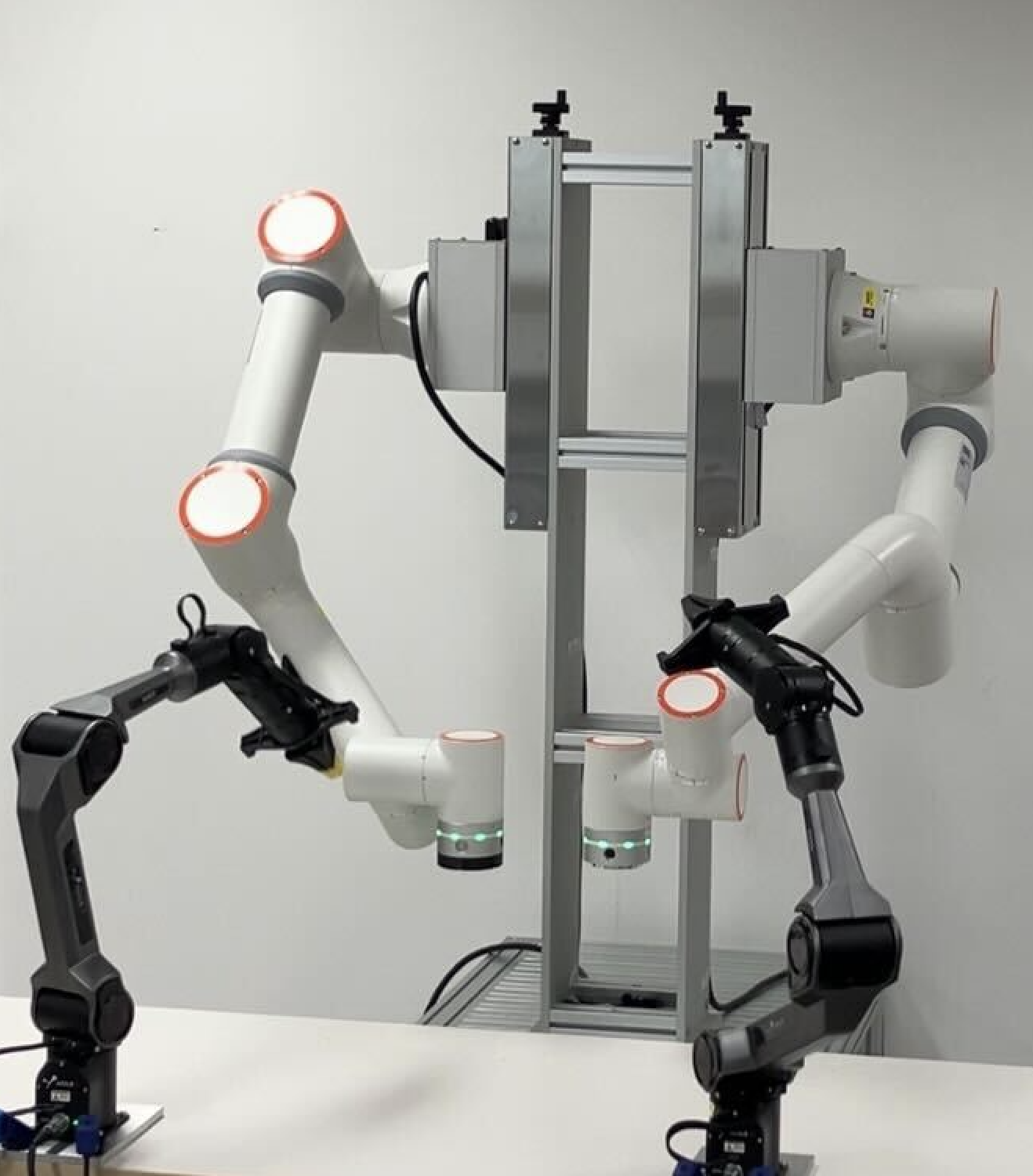}
    \hspace{0cm}
    \includegraphics[height=4cm]{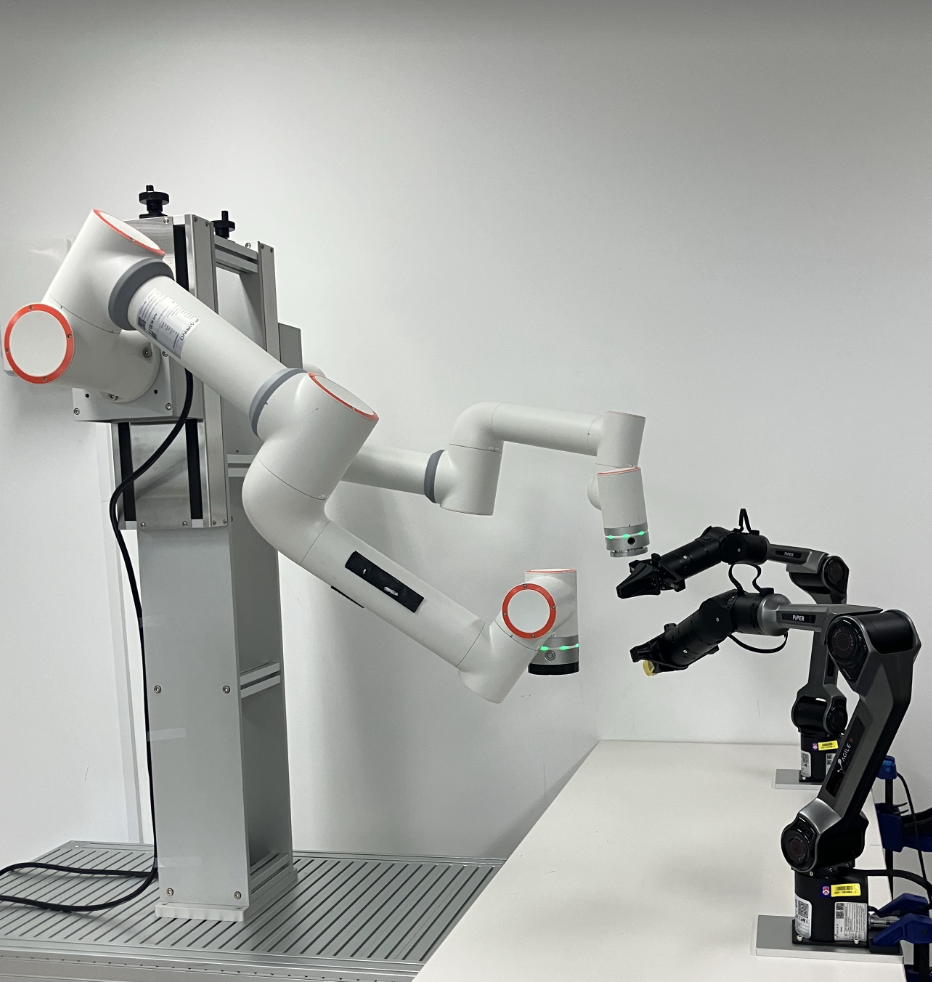} 
    \caption{A heterogeneous manipulator system with four 6-DOF robots}
    \label{fig:physical_exp}
    \vspace{-10pt}
\end{figure}

To validate the practical applicability of APF-ECBS, we used a physical multi-robot system composed of two FR5 manipulators from FAIRINO and two PiPER manipulators from Agile-X Robotics, both of which are 6-DOF. The manipulators were arranged to operate within a shared test area, with minimum and maximum distances between the robot bases of up to $1.5\textnormal{m}$ as shown in Fig.~\ref{fig:physical_exp}. We generated five distinct instances to test the algorithm's performance in various scenarios: two instances with end-effector crossing and parallel-pose trajectories, and three instances where the start and goal poses are similar to those in pick, place, and packing tasks. As shown in the supplementary video submission, our method can perform multi-manipulator coordination without any physical collisions.

\section{Conclusion}
\vspace{-3pt}

We proposed APF-ECBS, an MMP method designed to mitigate the inefficiency of cascading conflicts in CBS-based approaches. Its core is a repulsive trajectory modification at the low level, using an APF to create separation between manipulators and prevent future conflicts. At the high level, a fast-track strategy resolves one-to-many conflict patterns in a single step, accelerating the search. Experiments show that APF-ECBS achieves higher success rates and faster solutions than state-of-the-art algorithms, with significantly fewer expanded nodes. This efficiency comes at the cost of path quality, producing trajectories with larger makespan and cost, but making the method well suited for complex or time-critical scenarios where feasibility and speed outweigh optimality.


\clearpage
\bibliographystyle{IEEEtran}
\bibliography{references}

\end{document}